\documentclass{article}
\usepackage{arxiv}
\usepackage{graphicx}
\usepackage{hyperref}
\usepackage{url}
\usepackage{booktabs}
\usepackage{amsfonts}       
\usepackage{nicefrac}       
\usepackage{microtype}      
\usepackage{graphicx}
\usepackage{subfig}
\usepackage{color}
\usepackage{fancyhdr}

\makeatletter
\newcommand{\printfnsymbol}[1]{%
  \textsuperscript{\@fnsymbol{#1}}%
}
\makeatother

\begin{document}
\title{On Conditioning GANs to Hierarchical Ontologies
\thanks{This work was supported by Austrian Ministry for Transport, Innovation and Technology, the Ministry of Science, Research and Economy, and the Province of Upper Austria in the frame of the COMET center SCCH.}
}

\author{
 Hamid Eghbal-zadeh\thanks{Equal contribution.} \\
  LIT AI Lab \& Institute of Computational Perception\\
  Johannes Kepler University Linz, Austria\\
  \texttt{hamid.eghbal-zadeh@jku.at} \\
   \And
 Lukas Fischer\printfnsymbol{2}\\
  Software Competence Center Hagenberg GmbH (SCCH)\\
  Softwarepark 21, 4232 Hagenberg, Austria\\
  \texttt{lukas.fischer@scch.at} \\
  \And
 Thomas Hoch\printfnsymbol{2}\\
  Software Competence Center Hagenberg GmbH (SCCH)\\
  Softwarepark 21, 4232 Hagenberg, Austria\\
  \texttt{thomas.hoch@scch.at} \\
}

\maketitle

\begin{abstract}
The recent success of Generative Adversarial Networks (GAN) is a result of their ability to generate high quality images from a latent vector space. 
An important application is the generation of images from a text description, where the text description is encoded and further used in the conditioning of the generated image.
Thus the generative network has to additionally learn a mapping from the text latent vector space to a highly complex and multi-modal image data distribution, which makes the training of such models challenging. To handle the complexities of fashion image and meta data, we propose Ontology Generative Adversarial Networks (O-GANs) for fashion image synthesis that is conditioned on an hierarchical fashion ontology in order to improve the image generation fidelity. 
We show that the incorporation of the ontology leads to better image quality as measured by Fr\'{e}chet Inception Distance and Inception Score.
Additionally, we show that the O-GAN achieves better conditioning results evaluated by implicit similarity between the text and the generated image.

\end{abstract}

\section{Introduction}
\begin{figure}[ht]
\begin{center}
\begin{tabular}{cc}
\subfloat[]{\label{fig:ontology}\includegraphics[height=0.45\textwidth]{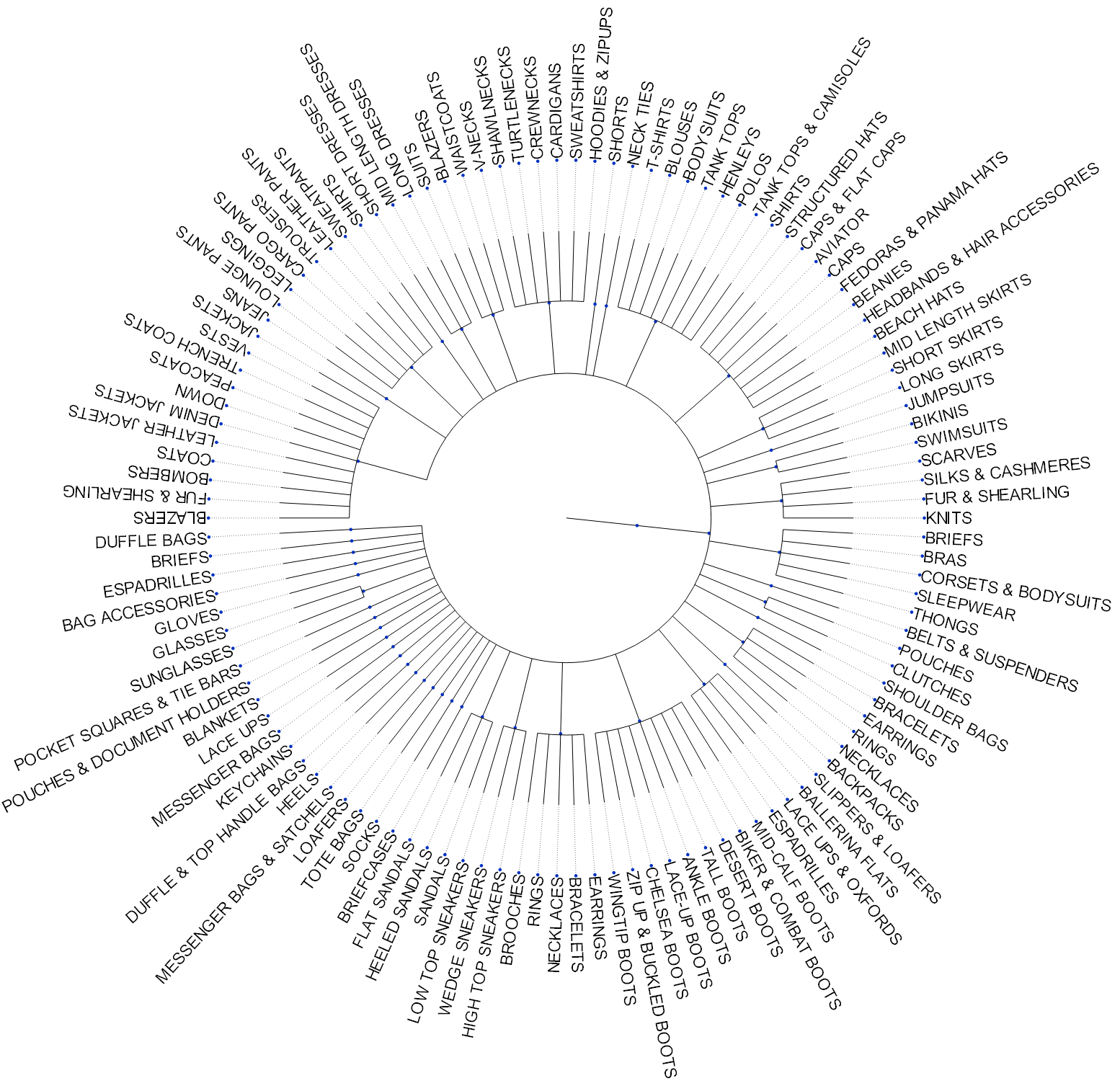}}%
\qquad
\subfloat[]{\label{fig:glovec_emb}\includegraphics[height=0.35\textwidth]{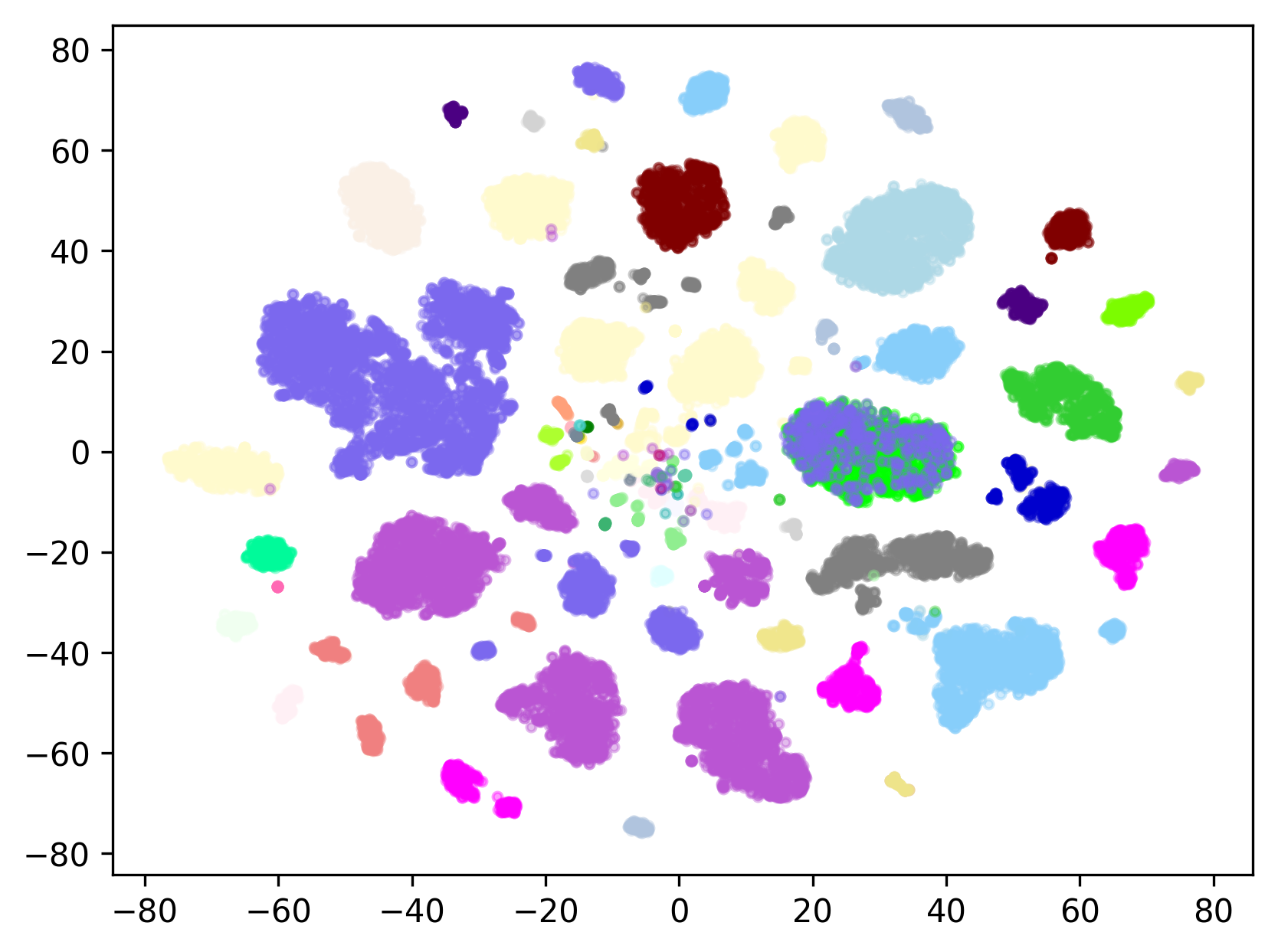}}
\end{tabular}
\caption{Fashion-Gen dataset: a) Ontology of sub-categories. b) Sub-category glovector embedding.}
\label{fig:subcats}
\end{center}
\end{figure}

Text-to-image synthesis is a challenging task where the details about the respective images are provided in a text corpus. The visual image details should best fit to the explanation provided in the text description, while maintaining a high-level of image detail fidelity. 
Generative adversarial networks (GANs) have proven to be a very powerful method to tackle this task ~\cite{gulrajani2017improved,karras_progressive_2018}. In \cite{Kuang2018} it has been shown that hierarchical model training by means of an ontology helps to learn more discriminant high-level features for fashion image representations. We adopt their strategy for generative models and show in this paper how a two layer fashion category ontology can be leveraged to improve GANs training for fashion image generation from text.
The recently organized Fashion-Gen challenge~\footnote[1]{https://fashion-gen.com/}~\cite{rostamzadeh_fashion-gen:_2018} provides a perfect test bed for the evaluation of novel methods for text-to-image synthesis. The provided dataset consists of $293.008$ images, with 48 main and 132 fine-grained categories as well as a detailed  description text. An ontology of sub-categories is visualized in Figure~\ref{fig:ontology}.
To handle the complexities of the Fashion-Gen challenge, we propose Ontology Generative Adversarial Networks (O-GANs) for high-resolution text-conditional fashion image synthesis. We detail the O-GAN in the following section.

\section{Ontology Generative Adversarial Networks}
\label{sec:gpgan}
Recently a new training methodology for GANs, namely progressive growing of GANs (PGAN~\cite{karras_progressive_2018}) was proposed to improve the variation, stability and quality for image generation. 
We introduce a modified and improved version of PGAN to cope with the challenges of high-resolution text-conditional image synthesis, called Ontology Generative Adversarial Networks (O-GANs).
A block-diagram of our proposed method is provided in Figure~\ref{fig:diag}.
We incorporate a progressive training based on PGAN and use a Wasserstein objective with gradient penalty~\cite{gulrajani2017improved} during training. 
For modelling the text, we use a word-level vector representation known as glovectors~\cite{pennington2014glove}. 
An embedding of the created sub-category glovectors can be found in Figure~\ref{fig:glovec_emb}. The utilized Discriminator and Generator are described in the following paragraphs:

\begin{figure}[ht]
\begin{center}
\includegraphics[width=\textwidth]{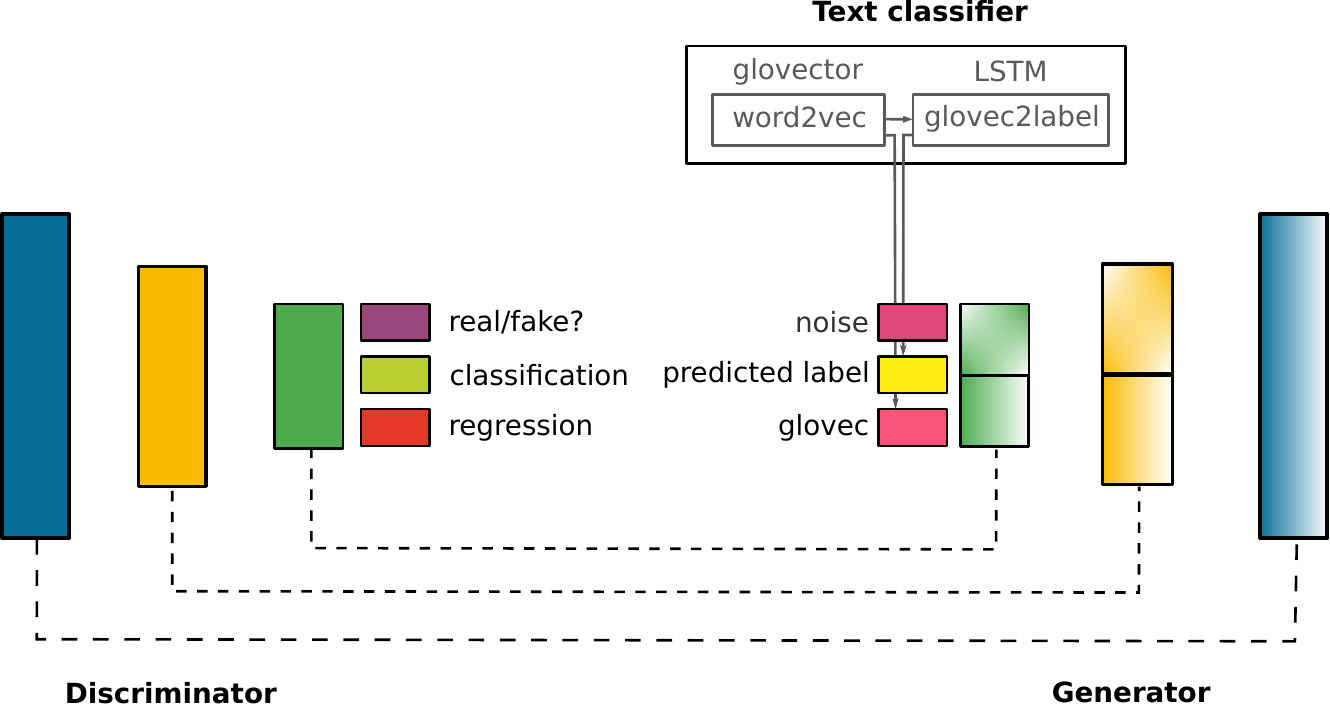}
\caption{Diagram of the proposed O-GAN.}
\label{fig:diag}
\end{center}
\end{figure}

Our model consists of 4 modules: 1) the Generator network $G$ to generate images, 2) the Discriminator network $D$ to distinguish between real and fake images, 3) the label predictor network $L$ to classify images and 4) the Regressor $R$ that regresses an image to its text embedding.
The discriminator $D$, label predictor $L$ and the Regressor $R$ share most of their model parameters and only have different output layers.

\textbf{The Discriminator $D$} was constructed having three outputs each minimizing one of three respective objectives.
This setting allows for parameter sharing between the $D$, $L$ and $R$ networks used for different objectives.
The first objective, is to minimize the Wasserstein distance between real and fake (generated) images in the $D$ network.
The second objective, is a \emph{classification} objective that minimizes the categorical cross entropy between labels (sub-categories) provided with the fashion images and the classification output in the $L$ network. 
Finally the third objective, is a \emph{regression} objective that minimizes an $L_2$ loss between the regression output of the $R$ network and the average of the glovectors of the words in the description provided with the image. 

\textbf{The Generator $G$} uses three concatenated vectors as the input: first the average glovectors of the text, second the labels as one-hot encoding and third a uniform random noise.
During generation time, a bidirectional LSTM~\cite{hochreiter1997long} trained on the sequences of glovectors is used to predict the labels from text.

\section{Evaluation Measures}
We use three evaluation measures to compare the proposed method with the PGAN baseline.
We evaluate the quality of the generated images from different models based on the Fr\'{e}chet Inception Distance (FID)~\cite{heusel2017gans} and Inception Score (IS)~\cite{salimans_improved_2016} (computed on 1k generated and 1k real images). The Inception model was trained on ImageNet~\cite{imagenet_cvpr09}.

To evaluate the quality of the conditioning, we report the cross-entropy between the conditioning labels and the probability of the labels for generated images, estimated via the label predictor $L$.
We train the label predictor $L$ to minimize the cross-entropy between the labels and the generated images through the probabilities produced in its output.
In addition, we report the $L_2$ distance between the glovector extracted from the conditioning text, and an estimated glovector via the Regressor model $R$.
The Regressor model $R$ also shares its weights with the discriminator, but has a separate output layer trained to minimize the $L_2$ distance between images and their corresponding glovector.
Hence, the Regressor model $R$ implicitly estimates the similarity between the generated image and the glovector used for the conditioning~\cite{chen2016infogan}.

\section{Results and Discussion}

\begin{figure}[ht]
\begin{center}
\begin{tabular}{cc}
\subfloat[]{\label{fig:fake_q_penalty_pgan}\includegraphics[width=.33\textwidth]{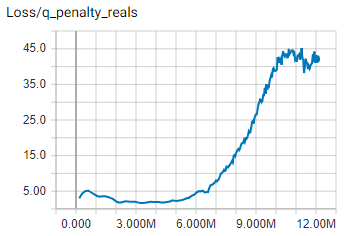}}
\subfloat[]{\label{fig:fake_q_penalty_tcpgan}\includegraphics[width=.33\textwidth]{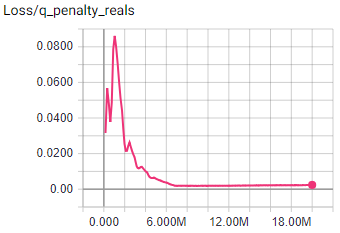}}
\subfloat[]{\label{fig:fake_label_penalty_pgan}\includegraphics[width=.33\textwidth]{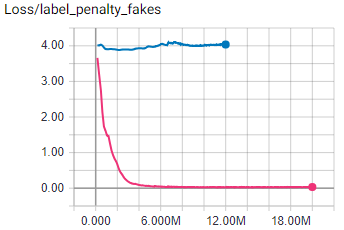}}
\end{tabular}
\caption{Comparison of discriminator losses for the modified PGAN (blue) and the proposed O-GAN (magenta).
a) $L_2$ distance between the condition glovector and the estimated glovector from images in PGAN by the Regressor model $R$ in different epochs.
b) $L_2$ distance between the condition glovector and the estimated glovector from images in O-GAN by the Regressor model $R$ in different epochs.
c) Cross-Entropy between label condition and the probabilities of the label predictor $L$ from the generated images in PGAN and O-GAN in different epochs.
}
\label{fig:cond_eval}
\end{center}
\end{figure}

\begin{figure}[ht]
\begin{center}
\begin{tabular}{cc}
\subfloat[]{\label{fig:fakes}\includegraphics[height=0.4\textwidth]{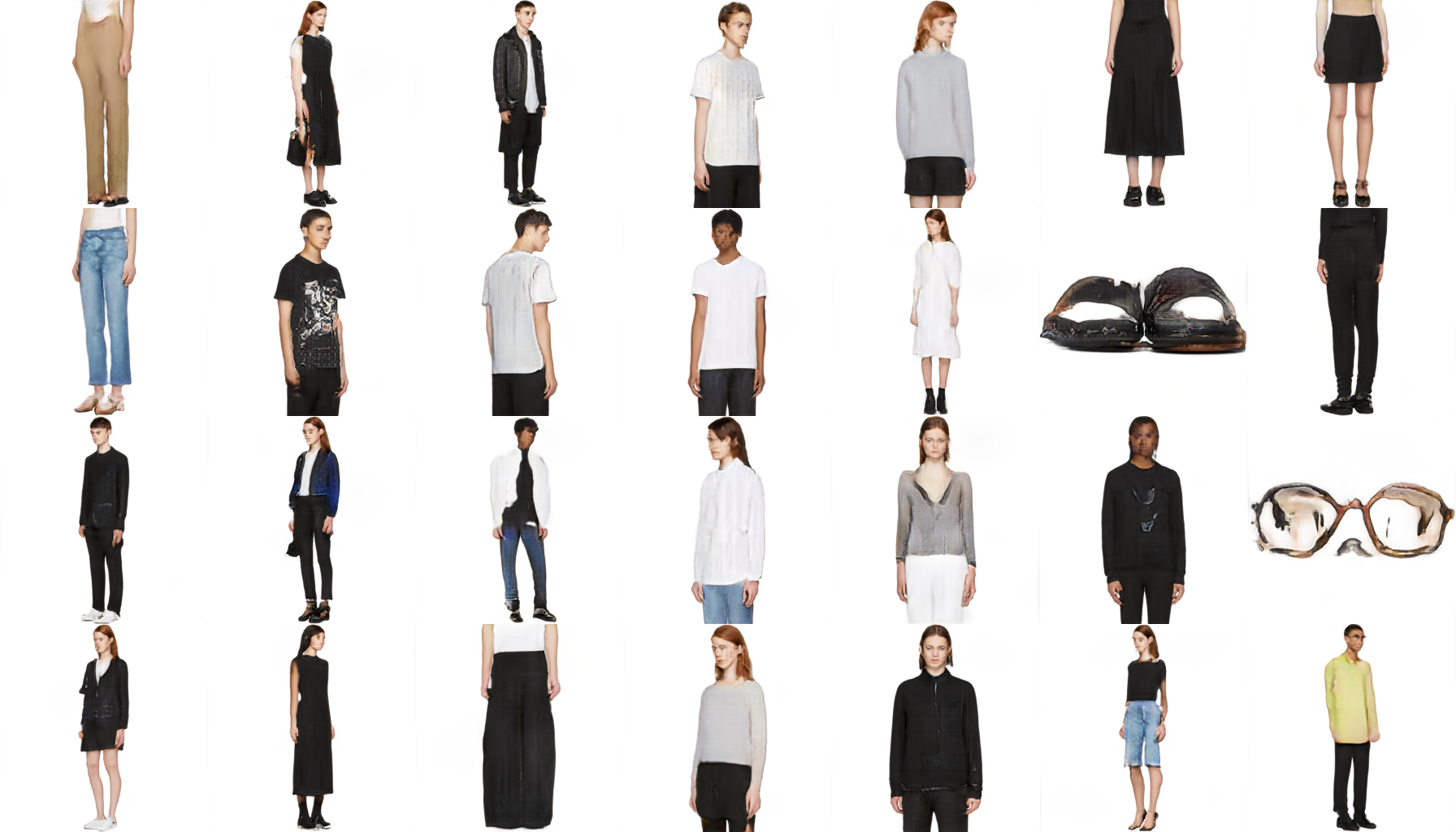}}\\
\subfloat[]{\label{fig:skirt}\includegraphics[height=0.35\textwidth]{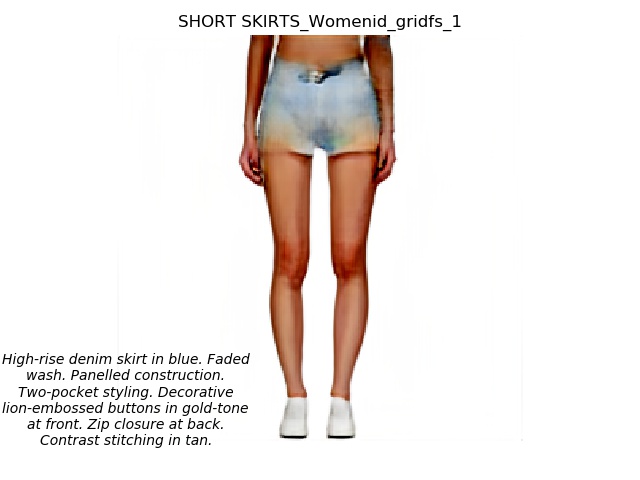}} 
\subfloat[]{\label{fig:bombers}\includegraphics[height=0.35\textwidth]{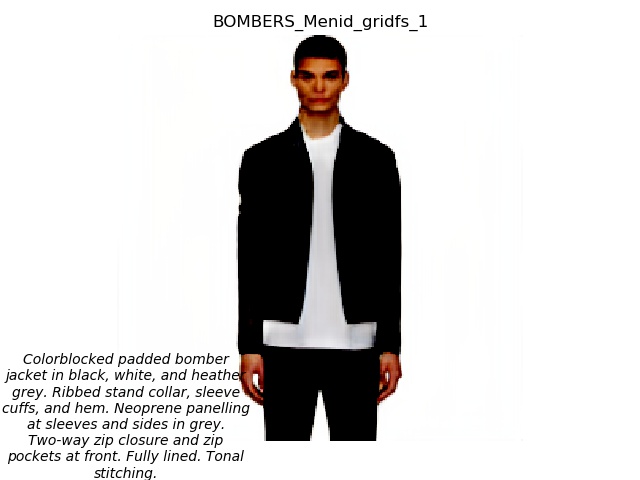}}\\
\end{tabular}
\caption{a) Random generated example images of our model. b-c) Generated images from our model and their conditioned text.}
\label{fig:all}
\end{center}
\end{figure}

In Table~\ref{eval_results}, we provide evaluation metrics on the Fashion-Gen dataset. Generated images of the implemented O-GAN are shown in Figure~\ref{fig:all}, where \ref{fig:fakes} depicts random generated examples and \ref{fig:skirt}-\ref{fig:bombers} show generated images with their corresponding textual description.
We used a modified PGAN as a baseline, in which a label predictor is added in addition to PGAN's discriminator. To demonstrate the importance of the ontologies, we only trained this baseline with category labels which has no information about the ontology of the finer-level classes.
Additionally the generator was modified to use the label as conditioning vector.
As shown in Figure~\ref{fig:cond_eval}, we can see that the O-GAN achieves lower $L_2$ distance to the conditioning glovector, compared to PGAN.
This suggests that the images generated by O-GAN have higher similarity to the conditioning text compared to PGAN, as estimated by the Regressor model $R$.
It can also be seen that a lower cross-entropy between the labels and the output of the label Regressor $R$ can be achieved which demonstrates better ability in label conditioning in O-GAN compared to PGAN.
As can be seen in Table~\ref{eval_results} and Figure~\ref{fig:cond_eval}, the proposed method outperforms the PGAN in all cases of the reported evaluations.
Training and evaluation was done on a NVidia DGX Station with 4 Tesla V100 GPUs using the Tensorflow framework.

\begin{table}
\begin{center}
\begin{tabular}{c|cc}
\hline
method \textbackslash measure & IS (1k)$\uparrow$ & FID (1k)$\downarrow$ \\ \hline
 Real Images & 5.03 $\pm$ 0.36 & 0 \\ \hline
 Vanilla PGAN & 4.54 $\pm$ 0.28 &  33.80  \\
 Proposed O-GAN & \textbf{4.81 $\pm$ 0.61} & \textbf{31.14} \\ \hline 

\end{tabular}%
\end{center}
\caption{Image quality evaluation results. For Inception Score (IS) higher values and for Fr\'{e}chet Inception Distance (FID) lower values are better.}
\label{eval_results}
\end{table}

\section{Conclusion}
In this paper, we proposed a novel Text-Conditional GAN, the O-GAN, capable of generating realistic high-resolution images from a text describing the characteristic of the target image. 
We showed that our GAN can be used for generating fashion images from text description provided by the experts in fashion industry.
We demonstrated the ability of O-GAN in incorporating ontologies in the generative process and showed how it improves the performance of both conditioning and quality of the generated images.
We also showed that the proposed model outperforms the PGAN model in terms of text-conditioning evaluation measures.
\bibliographystyle{plain}
\bibliography{refs}

\end{document}